\documentclass[11pt]{article} 

\usepackage[utf8]{inputenc} 

\usepackage{geometry} 
\usepackage{graphicx} 

\usepackage{booktabs} 
\usepackage{array} 
\usepackage{paralist} 
\usepackage{verbatim} 
\usepackage{subfig} 

\usepackage{fancyhdr} 
\usepackage{sectsty}
\allsectionsfont{\sffamily\mdseries\upshape} 

\usepackage[nottoc,notlof,notlot]{tocbibind} 
\usepackage[titles,subfigure]{tocloft} 

\usepackage{hyperref}

\def\metainfo#1{\textbf{#1}}

\def\metainfo#1{}

\title{Human-in-the-loop Artificial Intelligence}
\author{Fabio Massimo Zanzotto\\University of Rome Tor Vergata\\\small{\texttt{fabio.massimo.zanzotto@uniroma2.it}}}
\date{} 

\begin{document}
\maketitle

\begin{abstract}
Little by little, newspapers are revealing the bright future that Artificial Intelligence (AI) is building. Intelligent machines will help everywhere. However, this bright future has a dark side: a dramatic job market contraction before its unpredictable transformation. Hence, in a near future, large numbers of job seekers will need financial support while catching up with these novel unpredictable jobs. 
This possible job market crisis has an antidote inside. In fact, the rise of AI is sustained by the biggest knowledge theft of the recent years. Learning AI machines are extracting knowledge from unaware skilled or unskilled workers by analyzing their interactions. By passionately doing their jobs, these workers are digging their own graves. 

In this paper, we propose Human-in-the-loop Artificial Intelligence (HIT-AI) as a fairer paradigm for Artificial Intelligence systems. HIT-AI will reward aware and unaware knowledge producers with a different scheme: decisions of AI systems generating revenues will repay the legitimate owners of the knowledge used for taking those decisions. As modern Robin Hoods, HIT-AI researchers should fight for a fairer Artificial Intelligence that gives back what it steals. 

\end{abstract}

\section{Introduction}

\metainfo{Artificial Intelligence is Great}

We are on the edge of a wonderful revolution: Artificial Intelligence (AI) is breathing life into helpful machines, which will relieve us of our need to perform repetitive activities. Self-driving cars \cite{Driverless2016,lutin2013revolutionary,litman2014autonomous} are taking their first steps in our urban environment and their younger brothers, that is, assisted driving cars \cite{revathi2012smart,trajkovic2004computer,gray2012semi}, are already a commercial reality. Robots are vacuum cleaning and mopping the floors of our houses \cite{ulrich1997autonomous,taylor2005robot,huffman2008robotic}. Chatbots\footnote{see \url{http://houseofbots.com/}} \cite{weizenbaum1966eliza,Wallace2009} have conquered our new window-on-the-world --  our smartphones -- and, from there, they help with everyday tasks such as managing our agenda, answering our factoid questions or being our learning companions \cite{kerly2007bringing,beccaceci2009education}. In medicine, computers can already help in formulating diagnoses \cite{austin2013using,kourou2015machine,ferroni2017risk} by looking at data doctors generally neglect.    
Artificial Intelligence is preparing a wonderful future where people are released from the burden of repetitive jobs.

\metainfo{Artificial Intelligence is wiping out many jobs: mass unemployment}

The bright Artificial Intelligence revolution has a dark side: a dramatic mass unemployment that will precede an unpredictable job market transformation. People and, hence, governments are frightened. Nearly every week, newspapers all over the world are reporting on possible futures where around one fifth of actual jobs will disappear. Alarming reports foresee that more than one billion people will be unemployed worldwide \cite{McKinsey1}. 
By releasing people from repetitive jobs, intelligent machines will replace many workers. Chatbots are slowly replacing call center agents. Self-driving trains are already reducing the number of drivers in our trains. Self-driving cars are fighting to replace cab drivers in our cities. Drones are expanding automation in managing delivery of goods by drastically reducing the number of delivery people. And, these are only examples as even more cognitive and artistic jobs are challenged. Intelligent machines may produce music jingles for commercials \cite{briot2017deep}, write novels, produce news articles and so on. Intelligent risk predictors may replace doctors \cite{austin2013using,kourou2015machine,ferroni2017risk}. Chatbots along with massive open online courses may replace teachers and professors \cite{TEACHERBOTS}. Coders risk being replaced by machines too \cite{IBMendOfProgramming}. Nobody's job is safe as we face this overwhelming progress of Artificial Intelligence.

\metainfo{Massive stealing of competences and knowledge}

Surprisingly, the rise of Artificial Intelligence is supported by the unaware mass of people who risk seeing their jobs replaced by machines. These people are giving away their knowledge, which is used to train these wonderful machines. This is an enormous and legal knowledge theft taking place in our modern era. Along with those aware programmers and artificial intelligence researchers who set up the learning modules of these artificial intelligent machines, an unaware mass of people is providing precious training data by passionately doing their job or simply performing their activity on the net. Answering an email, an interaction on a messaging service, leaving an opinion on a hotel, and so on are all simple everyday activities people are doing. This data is a goldmine for artificial intelligence machines. Learning systems transform these interactions in knowledge for the artificial intelligence machines and the knowledge theft is completed. By doing their normal everyday activity, people are digging the grave for their own jobs. 

As researchers in Artificial Intelligence, we have a tremendous responsibility: building intelligent machines we can work with rather than intelligent machines that steal our knowledge to do our jobs. We need to find ways to financially support job seekers  as they  train to catch up with these novel unpredictable jobs. We need to prepare an antidote as we spread this poison in the job market.

This paper propose \emph{Human-in-the-loop Artificial Intelligence} (HIT-AI) as a novel paradigm for a responsible Artificial Intelligence. This is a possible antidote to the poisoning of the job market. The idea is simple: \emph{giving the right value to the knowledge producers}. Human-in-the-loop AI is an umbrella for researchers in Artificial Intelligence working with this underlying idea.
Hence,  HIT-AI promotes interpretable learning machines and, therefore, artificial intelligence systems with \emph{a clear knowledge lifecycle}. For HIT-AI systems, it will be clear whose the knowledge has been used in a specific deployment or in specific situations. This is a way to give the rightful credit and revenue to the original knowledge producers. We need a fairer artificial intelligence.

The rest of the paper is organized as follows. Section \ref{sec:enabling} describes the enabling paradigms of Human-in-the-loop AI. Section \ref{sec:proposals} sketches some simple proposals for a better future. Then, Section \ref{sec:conclusions} draws some conclusions.

\section{Human-in-the-loop AI: Enabling Paradigms}
\label{sec:enabling}

\subsection{Transferring Knowledge to Machines with Programming vs. with Learning from Repeated Experience}

Since the beginning of the digital era, \emph{programming} is the preferred way to \emph{``teach''}  to machines. \emph{Artificial non-ambiguous programming languages} have been developed to have a clear tool to tell machines what to do. 
According to this paradigm, whoever wants to \emph{``teach''} machines how to solve a new task or how to be useful has to master one of these programming languages. These people, called \emph{programmers}, have been \emph{teaching} machines for decades and have made these machines extremely useful. Nowadays, it is difficult to think staying a single day without using the big network of machines programmers have contributed to building.

As not all the tasks can be solved by \emph{programming}, \emph{autonomous learning} has been reinforced as an alternative way of controlling the \emph{``behavior''} of machines. In autonomous learning, machines are asked to \emph{learn from experience}. 
With the paradigm of \emph{programming},  we have asked machines to \emph{go to school} before these machines \emph{have learned to walk through trial and error}. This is why machines have always been good in solving very complex cognitive tasks but very poor in working with everyday simple problems. The paradigm of \emph{autonomous learning} has been introduced to solve this problem.

In these two paradigms, who should be paid for transferring knowledge to machines and how should they be paid? In the \emph{programming paradigm}, roles are clear: \emph{programmers} are the \emph{``teachers''} and \emph{machines} are the \emph{``learners''}. Hence, programmers could be payed for their work. In the \emph{autonomous learning paradigm}, the activity of programmers is confined to the selection of the most appropriate learning model and of the examples to show to these learning machines. 



From the point of view of HIT-AI, \emph{programming} is a fair paradigm as it keeps humans in the loop although machines, which have been taught exactly what to do, can hardly be called \emph{artificial intelligence}. On the contrary, \emph{autonomous learning} is an unfair model of transferring knowledge as the real knowledge is extracted from data produced by unaware people. Hence, little seems to be done by humans and machines seem to do the whole job. Yet, knowledge is stolen without paying.

\subsection{Explainable Artificial Intelligence and Explainable Machine Learning}

Explaining the decisions of learning machines is a very hot topic nowadays: dedicated workshops or specific sessions in major conferences are flourishing \cite{2017XAI,2017arXiv170802666K}. In specific areas of application, for example, medicine, thrust in intelligent machines cannot be blind as final decisions can have a deep impact on humans. Hence, understanding why a decision is taken become extremely important. However, what is exactly an explainable machine learning model is still an open debate \cite{DBLP:journals/corr/Lipton16a}. 

In HIT-AI, explainable machine learning can play a crucial role. In fact, seen from another perspective, explaining machine learning decisions can keep humans in the loop in two ways: 1) giving the last word to humans; and, 2) explaining what data sources are responsible for the final decision. In the first case, the decision power is left in the hand of very specialized professionals that use machines as advisers. This is a clear case of human-in-the-loop AI. Yet, this is confined to highly specialized knowledge workers in some specific area. The second case instead is fairly more important. In fact, machines that take decisions or work on a task are constantly using knowledge extracted from data. Spotting which data have been used for a specific decision or for a specific action of the machine is very important in order to give credits to who has produced these data. In general, data are produced by anyone and everyone, not only by knowledge workers. Hence, understanding why a machine takes a decision may become a way to keep everybody in the loop of artificial intelligence.

\subsection{Convergence between Symbolic and Distributed Knowledge Representation}

Explaining machine learning decisions is simpler in image analysis o, better, in all those cases where the system representation is similar what is represented. In fact, for example, neural networks interpreting images are generally interpreted by visualizing how subparts represent salient subparts of target images. Both input images and subparts are tensors of real numbers. Hence, these networks can be examined and understood.

However, large part of the knowledge is expressed with symbols. Both in natural and artificial languages, combination of symbols are used to convey knowledge. In fact, for natural languages, sounds are transformed in letters or ideograms and these symbols are composed to produce words. Words then form sentences and sentences form texts, discourses, dialogs, which ultimately convey knowledge, emotions, and so on. This composition of symbols into words and of words in sentences follow rules that both the hearer and the speaker know \cite{Chomsky1957}. Hence, symbolic representations give a clear tool to understand whose knowledge is used in specific machines.


In current Artificial Intelligence systems, symbols are fading away, erased by tensors \emph{distributed representations}. \emph{Distributed representations} are pushing deep learning models \cite{lecun2015deep,schmidhuber2015deep} towards amazing results in many high-level tasks such as  image recognition \cite{he2016identity,simonyan2014very}, image generation \cite{goodfellow2014generative} and image captioning \cite{vinyals2015show,xu2015show}, machine translation \cite{bahdanau2014neural,zou2013bilingual}, syntactic parsing \cite{NIPS2015_5635,weiss2015structured} and even game playing at human level \cite{silver2016mastering,mnih2013playing}. 


There is a strict link between distributed representations and symbols,  the first being an approximation of the second \cite{Plate:1994,Plate1995,ferrone-zanzotto:2014:Coling,DBLP:conf/slsp/FerroneZC15}.  The representation of the input and the output of these networks is not that different from their internal representation.

For HIT-AI, this strict link is a tremendous opportunity to track how symbolic knowledge flows in the knowledge lifecycle. In this way, symbolic knowledge producers can be rewarded for their unaware work.


\section{Human-in-the-loop AI: a simple proposal for a better Future}
\label{sec:proposals}

A peasant of the late 19th century would have never imagined that after 100 years \emph{yoga trainer}, \emph{pet caretaker} and \emph{ayurveda massage therapist} -- just to cite technology unrelated jobs -- are common jobs. It is also extremely likely that any wise politician of that period had the same lack of imagination even though s/he had more time to spend to imagine the future and less pressure on her/his job loss.

Today, we are in a situation similar to the end of the 19th century but we have a complication: the speed of the AI revolution. As it was for the end-of-19th-century peasants and politicians, we can hardly imagine what's next on the job market. We can see some trends but it is hard to exactly imagine what are the skills needed for being part of the labor force of the future. Yet, the AI revolution is overwhelming and risks elimination of many jobs in the near future. 
This may happen before our society envisage a clear path for relocating workers. We urge a strategy for the immediate.

The Artificial Intelligence revolution is based on an enormous knowledge theft. Skilled and unskilled workers do their own everyday jobs and leave important traces. These traces are the \emph{training examples} that machines can use to learn. Hence, Artificial Intelligence using machine learning is stealing these workers' knowledge by learning from their interactions. These unaware workers are basically digging the graves for their own jobs.

The knowledge produced by workers and used by machines is going to produce revenues for machine owners for years. This is a major problem since only a very small fraction of the population can benefit from this never-ending revenue source and the real owners of the knowledge are not participating to this redistribution of wealth.

The model we propose with Human-in-the-loop Artificial Intelligence seeks to give back part of the revenues to the unaware knowledge producers. 

The key idea is that any profit-making interaction a machine does has to constantly repay whoever has produced the original knowledge used to do that interaction. To obtain repayment, we need to work on a major issue: 
determine \textbf{a clear knowledge lifecycle} which performs a compete tracking of the knowledge from its initial production to the final decision processes of the machine.
Hence, we need to promote artificial intelligence models that are explainable and that track back to the initial training examples that originated a decision. In this way, it is clear why the decision is made and who has to be rewarded with a fraction of the profit that the decision is producing.   

Managing ownership in the knowledge life-cycle poses big technological and moral issues and it is certainly more complex than \emph{simply using} knowledge while forgetting what the source is. 
Each interaction has to be tracked and assigned to a specific individual. Hence, the issues are: first, a clear identification of people in the web is mandatory; second, privacy can become an overwhelming legal issue. 

Finally, to pursue HIT-AI as an ecosystem for fair artificial intelligence solutions, we need to invest in the following enabling technologies and legal aspects: 
\begin{itemize}
\item \emph{Explainable Artificial Intelligence} which is a must because, in order to reword knowledge producers, systems need to exactly know who is responsible for a specific decision;
\item \emph{Symbiotic Symbolic and Distributed Knowledge Representation Models} which are needed as a large part of knowledge is expressed with symbols;
\item \emph{Trusted Technologies} as the knowledge life-cycle should be clear and correctly tracked;
\item \emph{Virtual Identity Protocols and Mechanisms} because systems need exactly who has to be reworded;
\item \emph{Privacy Preserving Protocols and Mechanisms} as, although systems need to know who should be rewarded and why, privacy should be preserved;
\item \emph{Studying Extensions of Copyright to unaware knowledge production} which can be the legal solution to safeguard the unaware knowledge producers.
\end{itemize}

\section{Conclusions}
\label{sec:conclusions}

Job market contraction is the dark side of the shining future promised by Artificial Intelligence (AI) systems. Unaware skilled and unskilled knowledge workers are digging graves for their own jobs by passionately doing their normal, everyday work. Learning AI are extracting knowledge from their interactions. This is a gigantic knowledge theft of the modern era. 

In this paper, we proposed Human-in-the-loop Artificial Intelligence (HIT-AI) as a fairer AI approach. As modern Robin Hoods, HIT-AI researchers should fight for a fairer Artificial Intelligence that gives back what it steals. As skilled and unskilled workers are producing the knowledge which Artificial Intelligence is making profit on, we need to give back a large part of this profit to its legitimate owners. 

\newpage

\bibliographystyle{plain}
\bibliography{super,refined_bibliography}

\end{document}